\documentclass[9pt, a4paper]{extarticle}

\usepackage{charter} 
\usepackage[margin=.6in, top=1.75cm, bottom=2.4cm, headsep=0.2cm, footskip=1cm]{geometry}  % margin
\usepackage{multicol}
\usepackage{needspace}

\usepackage[utf8]{inputenc} 
\usepackage[T1]{fontenc} 
\usepackage[labelfont=bf, skip=5pt]{caption}
\usepackage{subcaption}

\usepackage{graphicx}
\graphicspath{{images/}}
\usepackage{dblfloatfix} % 'b' float place the figure at the bottom of the page
\newenvironment{figcol}
  {\par\medskip\noindent\minipage{\linewidth}}
  {\endminipage\par\medskip}

\usepackage{fancyhdr}
\fancypagestyle{footer}{ 
    \fancyhf{}
    
    \fancyfoot[C]{\textbf{\thepage}}
    \fancyfoot[L]{\today}}
\usepackage{eso-pic}
\newcommand\LowerRight[1]{\AtPageLowerLeft{%
\put(\LenToUnit{0.88\paperwidth},\LenToUnit{0.04\paperheight}){#1}}}

\usepackage{enumitem} 

\usepackage{breqn} % multiline equations
\usepackage{amsmath,amssymb,bbm,mathdots} % useful symbols
\usepackage{mathtools} % declare operators
\usepackage{relsize} % enlarge symbols
\usepackage[misc]{ifsym} % letter symbol

\usepackage[table, xcdraw, dvipsnames]{xcolor}
\usepackage{makecell}

\usepackage{tikz}
\usetikzlibrary{automata,arrows.meta,positioning,calc}
\newcommand{\thickuparrow}{%
    \tikz[baseline=0.2ex]{
        \draw[line width=1.5pt, -{Stealth[round, length=1.2ex, width=1.5ex]}] (0,0) -- (0,2ex);
    }%
}
\newcommand{\thickdownarrow}{%
    \tikz[baseline=0.2ex]{
        \draw[line width=1.5pt, -{Stealth[round, length=1.2ex, width=1.5ex]}] (0,2ex) -- (0,0);
    }%
}

\definecolor{myred}{HTML}{720414}
\definecolor{mygreen}{HTML}{055b2e}
\definecolor{myblue}{HTML}{1b038a}
\definecolor{coldef}{HTML}{474747}
\newcommand{\hlt}[1]{\textcolor{myred}{\textit{#1}}}
\newcommand{\better}[1]{\textcolor{mygreen}{\textbf{#1}}}
\newcommand{\worse}[1]{\textcolor{myred}{\textbf{#1}}}

\newcommand{\fdot}[2][black]{%
  \textcolor{#1}{\scalebox{#2}{$\bullet$}}%
}
\definecolor{Ccam}{rgb}{0.1904,0.5831,0.6936}
\definecolor{Cnaive}{rgb}{0.02635,0.19635,0.6035}
\definecolor{Csoft}{rgb}{0.21335,0.4896,0.2465}
\definecolor{Cpinet}{rgb}{0.11645,0.71995,0.03315}
\definecolor{Cfeedback}{rgb}{0.7803,0.74375,0.13685}
\definecolor{Censemble}{rgb}{0.78625,0.47685,0.13685}
\definecolor{Cstrong}{rgb}{0.8398,0.0935,0.13345}

\usepackage{titlesec}
\titleformat{\section}[block] % {command}[shape]
  {\sffamily\huge\bfseries\color{myred}} % {format}
  {\thesection} % {label}
  {0.5em} % {sep}
  {} % No {before-code}
  [\vspace{-0.5em}\color{myred}\rule{\linewidth}{2pt}] % {after-code}

\newcommand{\ptitle}[1]{%
    \phantomsection
    \hypertarget{ptitle:#1}{}%
    \noindent\textcolor{black}{\textbf{#1}}%
    \addcontentsline{toc}{subsection}{#1}%
    \normalsize
}

\usepackage{csquotes}

\usepackage[subfigure]{tocloft}

\usepackage[numbers,sort&compress]{natbib}
\usepackage[colorlinks = true,
            linkcolor = myblue,
            urlcolor  = myred,
            citecolor = myred,
            anchorcolor = myblue]{hyperref}
\usepackage{lipsum}

\usepackage[hang,flushmargin]{footmisc}

\usepackage{varwidth}
\usepackage[most]{tcolorbox}
\tcbuselibrary{skins}
\newtcolorbox{mybox}[2][]{
  enhanced, breakable,
  before skip=5mm,after skip=5mm,
  left=0.3cm, right=0.3cm,
  colback=black!5,colframe=black!50,boxrule=0mm,
  attach boxed title to top left={xshift=1cm,yshift*=1mm-\tcboxedtitleheight},
  varwidth boxed title*=-3cm,
  boxed title style={
    frame code={
      \path[fill=tcbcolback!30!black]
        ([yshift=-1mm,xshift=-1mm]frame.north west)
        arc[start angle=0,end angle=180,radius=1mm]
        ([yshift=-1mm,xshift=1mm]frame.north east)
        arc[start angle=180,end angle=0,radius=1mm];
      \path[left color=tcbcolback!60!black,right color=tcbcolback!60!black,
            middle color=tcbcolback!80!black]
        ([xshift=-2mm]frame.north west) -- ([xshift=2mm]frame.north east)
        [rounded corners=1mm]--
        ([xshift=1mm,yshift=-1mm]frame.north east) -- (frame.south east) --
        (frame.south west) -- ([xshift=-1mm,yshift=-1mm]frame.north west)
        [sharp corners]-- cycle;
    },
    interior engine=empty,
  },
  title={#2},#1
}

\usepackage[strict]{changepage}
\usepackage{framed}
\definecolor{formalshade}{rgb}{0.98,0.98,0.98}
\newenvironment{formal}[1][0.48]{%
  \MakeFramed{\hsize#1\textwidth\advance\hsize-1\width\FrameRestore}%
  \noindent\hspace{-4.55pt}% disable indenting first paragraph
  \begin{adjustwidth}{}{7pt}
  \vspace{8pt} % Add vertical space above the text
}
{%
  \vspace{5pt}\end{adjustwidth}\endMakeFramed%
}

\newcommand{\X}{\mathcal X}
\newcommand{\Y}{\mathcal Y}
\newcommand{\Z}{\mathcal Z}

\newcommand{\x}{\mathbf{x}}
\newcommand{\y}{\mathbf{\hat y}}
\newcommand{\z}{\mathbf{z}}

\newcommand{\h}{\mathbf{\hat h}}
\newcommand{\pibold}{\boldsymbol{\pi}}
\newcommand{\piboldhat}{\boldsymbol{\hat\pi}}
\newcommand{\pib}{\boldsymbol{\hat\pi}(\x)}
\newcommand{\pibk}{\boldsymbol{\hat\pi}_k(\x)}
\newcommand{\pibh}{\boldsymbol{\tilde\pi}(\x)}
\newcommand{\pibs}{\boldsymbol{\pi}(\x)}
\newcommand{\parent}[1]{\left( #1 \right)}

\newcommand{\set}[1]{\left\{ #1 \right\}}

\begin{document}
%%%%%%%%%%%%%%%%

%%%%%%%%%%%%%%%%%%%%%%%%%%%%%%%%%%%%%%%%%%%%%%%%%%%%%%%%%%%%%
%%%%%%%%%%               Cover Page               %%%%%%%%%%%
%%%%%%%%%%%%%%%%%%%%%%%%%%%%%%%%%%%%%%%%%%%%%%%%%%%%%%%%%%%%%

% \setcitestyle{round}
\captionsetup{width=.9\linewidth, font={small}}

\AddToShipoutPictureFG{
  \LowerRight{{\href{https://quasimodel.com/}{\includegraphics[width=1cm]{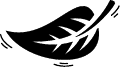}}}}%
}%
\pagestyle{footer}

%% reduce space after section header
\titlespacing*{\section}{0pt}{1.5ex plus 0.5ex minus 0.2ex}{0.8ex plus 0.2ex}

\begin{center}
  
  {\color{myred}\rule{\textwidth}{.15cm}}\\[8pt]
  \Huge{\textbf{Aligned explanations in neural networks}} \\ \vspace{-4pt}
  {\color{myred}\rule{.75\textwidth}{.1cm}}\\[8pt]
  
  % \huge{\textcolor{black!50}{Author} {1}\textsuperscript{{ \small1, \Letter}}
  % \qquad \textcolor{black!50}{Author} {2}\textsuperscript{{ \small1, 2}}} \\ 
  
  \huge{\textcolor{black!50}{Corentin} {Lobet}\textsuperscript{{ \small1, \Letter}}
  \qquad \textcolor{black!50}{Francesca} {Chiaromonte}\textsuperscript{{ \small1, 2}}} \\ 
  
  {\color{black}\rule{\textwidth}{.05cm}}\\ 
    
\end{center}

\begin{center}

  % \noindent\large
  % \textsuperscript{1} Affiliation 1 \\[3pt]
  % \textsuperscript{2} Affiliation 2 \\[5pt]
  % \raisebox{-.2ex}{\Letter} $\ $ \href{mailto:author.1@gmail.com}{author.1@gmail.com} $\quad$
  
  \noindent\large
  \textsuperscript{1} Institute of Economics and L'EMbEDS, Sant'Anna School of Advanced Studies - Italy \\[3pt]
  \textsuperscript{2} Dept. of Statistics and Huck Institutes of the Life Sciences, Penn State University - USA \\[5pt]
  \raisebox{-.2ex}{\Letter} $\ $ \href{mailto:corentin.lobet@santannapisa.it}{corentin.lobet@santannapisa.it} $\quad$
  \raisebox{-1ex}{\includegraphics[height=0.6cm]{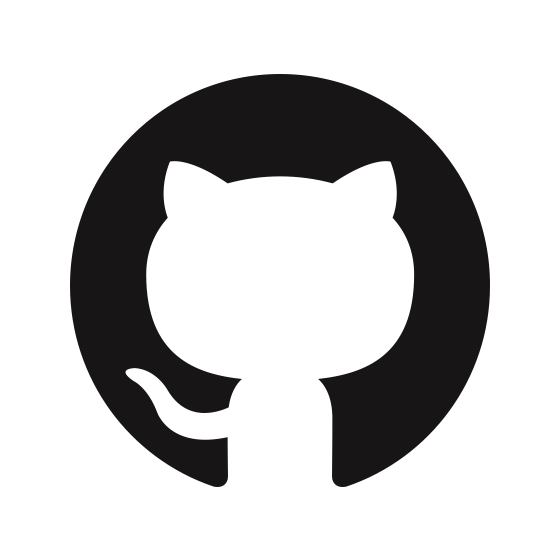}} $\ $ \href{https://github.com/FractalySyn/PiNets-Alignment}{Paper Code}
  \normalsize
  
  \vspace{0.15cm}
  {\color{black}\rule{\textwidth}{.05cm}}
  
\end{center}

% \vspace{.5cm}

\begin{mybox}[colbacktitle=black]{\large{Abstract}}

    \large
    As artificial intelligence increasingly drives critical decisions, the ability to genuinely explain how neural networks make predictions is essential for trust. Yet, most current explanation methods offer post-hoc rationalizations rather than guaranteeing a true reflection of the model's reasoning. We introduce the notion of explanatory alignment, a requirement that explanations directly construct predictions rather than rationalize them. To achieve this in complex data domains, we present Pointwise-interpretable Networks (PiNets), a pseudo-linear architecture that forms linear models instance-wise. Evaluated on image classification and segmentation tasks, PiNets demonstrate that their explanations are deeply faithful across four criteria: meaningfulness, alignment, robustness, and sufficiency (MARS). Our contributions pave the way for promising avenues: by reconciling the predictive power of deep learning with the interpretability of linear models, PiNets provide a principled foundation for trustworthy AI and data-driven scientific discovery.
    \normalsize

    % \hfill\\
    % \normalsize{ \textbf{Field Keywords:} Deep Learning, Explainable AI, Feature Attribution, Segmentation, Faithfulness } \\
    % \normalsize{ \textbf{Novel Keywords:} Explanatory Alignment, Model Readability, Pseudo-linearity, PiNet }
\end{mybox}

%%%%%%%%%%%%%%%%%%%%%%%%%%%%%%%%%%%%%%%%%%%%%%%%%%%%%%%%%%%%%%%%%%%%%%%
%%%%%%%%%%%%%%%%%%%%%%%%%%%%%%%%%%%%%%%%%%%%%%%%%%%%%%%%%%%%%%%%%%%%%%%
%%%%%%%%%%%%%%%%%%%%%%%%%%%%%%%%%%%%%%%%%%%%%%%%%%%%%%%%%%%%%%%%%%%%%%%

\begin{multicols}{2}
\captionsetup{width=.9\linewidth, font={small}}
\setlength{\parskip}{0pt}

  \ptitle{The misalignment problem}     
  The trust we place in AI-driven decisions increasingly depends on the quality of their justifications. As machine learning models penetrate high-stakes domains — from medical diagnosis and scientific discovery to autonomous systems and policy-making — the ability to explain how predictions are constructed becomes essential. Yet many existing explanation methods offer little guarantee that their explanations reflect the model's actual prediction-making process. When a decision is justified by a post-hoc rationalization or an ambiguous explanation, its trustworthiness is fundamentally compromised.

  Consider a neural network trained to detect cancer from medical images. If the model bases its prediction on irrelevant artifacts in the scan but post-hoc explanations highlight plausible tissue regions, clinicians may trust a prediction for the wrong reasons. The explanation becomes what we call \hlt{misaligned} — disconnected from how the model truly operates. Cognitive shortcuts are well-documented in human reasoning \citep{DISSONANCE,MORETHANWEKNOW,HEURISTICS,CHOICE} and we have both the opportunity and the responsibility to design AI systems that avoid them by construction \citep{MYTHOS,STOP}.

  Feature attribution — assigning importance scores to a set of features — is a major paradigm in explainable AI research \citep{SURVEY1,SURVEY2,SURVEY3,SURVEY4}. For a model producing prediction $\hat y\in\Y$ from input $\x\in\X$ (i.e., $\hat y = \hat f(\x)$), we seek an explanation $\pibh\in\Z$ that attributes scores to features $\z\in\Z$ ($\X=\Z$ is a common choice but not mandatory). Many model-agnostic methods \citep{SHAP, LIME, RISE, ANCHORS, L2X} like SHAP and LIME estimate these attributions by querying the model in the vicinity of $\x$ and inferring feature importance from the resulting local perturbations. However, this estimation process introduces fundamental ambiguities. When features are correlated, multiple explanations can fit the same local behavior equally well — yet only one (if any) corresponds to the model's true reasoning. The computational cost of resolving this ambiguity through exhaustive perturbation is often prohibitive, widening the gap between theory and practice \citep{PERTURB1,PERTURB2}.

  More subtly, model-agnostic approaches presuppose that the model has intrinsic attributions $\pib$ to discover — they produce an estimation $\pibh$ of $\pib$. But most models are not designed to produce feature attributions intrinsically in the first place, or may do so in a different feature space $\Z$. Any post-hoc explanation is necessarily a rationalization rather than a reflection of the model's internal logic, and the chances that the rationalization lands on the intrinsic explanation, if it exists, are slim.

  \hfill\\
  \ptitle{Aligned explanations and readability} 
  Intrinsic explanation methods avoid this rationalization trap. Yet intrinsicness alone proves insufficient.

  Gradient-based methods \citep{VANILLA,GRADCAM,INTGRAD,GRADEVAL1,GRADEVAL2,GRADEVAL3} compute attributions from the model's gradients, but gradients do not construct predictions — they are byproducts of internal computations rather than building blocks of predictions. Their interpretation as feature importance is therefore ambiguous.

  Other approaches to intrinsic explainability build feature attributions within the neural network architecture, either in parallel with predictions \citep{LMAC} or prior to them \citep{INVASE, REALX, COMET, BCOS, P2P}. Parallel attributions do not construct predictions directly and thus produce rationalizations subject to the same caveats as post-hoc methods. Prior attribution methods are more promising, but their usefulness depends the computational proximity of explanations to predictions. If attribution scores undergo complex transformations before forming predictions, their interpretation as explanations is limited.

  Concept-based models \citep{CONCEPT1,CONCEPT2,CONCEPT3} enhance traditional prediction networks with added interpretability over the internal features $\h(\x)$ produced by the model. They can provide attributions that directly precede predictions, but their interpretability is undermined by subjectivity in labeling learned features \citep{CONCEPTFAIL1,CONCEPTFAIL2}. If the model cannot produce unambiguous concept features, it becomes difficult to trust the explanations built upon them.

  Alignment requires more than mere intrinsicness. An explanation must not simply reflect a computationally-distant fragment of the prediction-making process; it must clearly reveal how the prediction is constructed from interpretable features. We propose \hlt{explanatory alignment} as the requirement that explanations genuinely underlie predictions rather than rationalize them after the fact. Achieving alignment demands three conditions jointly: attributions must be \hlt{intrinsic} to the model, exhibit \hlt{immediate precedence} (directly constructing predictions through simple operations), and operate on \hlt{fully interpretable} features.

  Alignment is fundamentally tied to model architecture and \hlt{model readability} is a key principle through which alignment translates into practice. We deem a model readable if it can be expressed as $\hat{y} = g(\pib, \z)$, where $\pib$ is intrinsic, $\z$ comprises only interpretable features, and $g$ is a simple aggregation function. We formalize these concepts in the Methods.

  Linear models exemplify readability — predictions emerge from weighted feature combinations. Yet their predictive power is rather limited; in particular in non-tabular data settings. Neural networks, in contrast, can extract information from complex data structures at the expense of interpretability. Reconciling the two approaches stands out as a promising path toward models that effectively combine predictive and explanatory power. 

  \hfill\\
  \ptitle{Pseudo-linear networks} 
  Rather than learning complex internal features $\h(\x)$ and combining them linearly as standard neural networks do, pseudo-linear models learn varying coefficients $\pib$ that linearly combine user-specified interpretable features $\z$:
  \begin{equation}
    \hat{y} = \hat a + \sum\nolimits_* \pib \circ \z
  \end{equation}
  where $\circ$ denotes element-wise multiplication and $\sum_*$ element-wise summation. This architecture constructs linear models instance-wise — each input $\x\in\X$ gives rise to a unique set of linear coefficients.

  The mechanism enabling this is what we call the \hlt{second look}. The coefficients $\pib \in \Z$ are multiplied element-wise by the features $\z \in \Z$. The model examines the data twice: first through $\x$ and then through $\z$. Importantly, $\X$ and $\Z$ can be identical, overlapping, or distinct in nature. 
  
  When $\Z$ is interpretable, alignment is satisfied since the coefficients $\pib$ precede and construct predictions. 

  For complex, non-tabular data like images, audio, or genomic sequences, neural networks provide the expressiveness needed for $\pib$ to capture intricate instance-specific patterns while preserving linear readability. 
  
  However, the second look alone is insufficient. While we argue that misaligned explanations are difficult to trust regardless of how convincing they look, aligned explanations are equally unuseful if they highlight spurious or random-looking signals in the data. 

  SENNs \citep{SENN,SURVEY3} pioneered pseudo-linear networks but with a major caveat: the recourse to concept learning to build the interpretable feature space. This approach constructs predictions by combining the outputs of two black boxes: the learned \enquote{concept} features $\h(\x)$ (in place of $\z$) and the varying coefficients $\pib$. Traditional neural networks already combine learned features linearly in their final layer, and thus on a single black box to construct predictions. SENNs' added value is therefore questionable: they introduce a second black box (varying coefficients) in a framework where interpretability is already hindered by the black box producing features. Alignment easily breaks in these settings.
  
  \hfill\\
  \ptitle{Contributions} 
  We address these limitations through a principled architectural approach. We present PiNets, a modeling framework for designing pseudo-linear neural networks that can produce aligned and meaningful explanations in non-tabular datasets. PiNets shift the black box from internal features to coefficients so that the explainability exercise now lies in understanding how the model combines interpretable features rather than in \enquote{conceptualizing} internal features.

  We also propose the MARS framework, a set of four principled criteria over explanatory faithfulness (including alignment). Through careful architectural design and training techniques — recursive stabilization, ensembling, and strong supervision — PiNets are shown to achieve high faithfulness across MARS criteria without compromising predictive accuracy. 

  In particular, PiNets are shown to match or exceed the performance of Grad-CAMs \citep{GRADCAM} in explaining image classifications while guaranteeing alignment and requiring less fine-tuning effort. We demonstrate that carefully designed PiNets are constrained to structure meaningful explanations out of statistical intelligence — acquired through training — as a precondition for predictive accuracy. 

  We showcase one implication of this property: PiNets can perform semantic segmentation by regressing image-level numerical targets while approaching the sharpness of models trained with pixel-level information. 

\end{multicols}

%%%%%%%%%%%%%%%%%%%%%%%%%%%%%%%%%%%%%%%%%%%%%%%%%%%%%%%%%%%%%%%%%%%%%%%
%%%%%%%%%%%%%%%%%%%%%%%%%%%%%%%%%%%%%%%%%%%%%%%%%%%%%%%%%%%%%%%%%%%%%%%
%%%%%%%%%%%%%%%%%%%%%%%%%%%%%%%%%%%%%%%%%%%%%%%%%%%%%%%%%%%%%%%%%%%%%%%

\needspace{3cm}
\section{Results}
\label{sec:sec2}

\begin{multicols}{2}
\setlength{\parskip}{0pt}

    \begin{figure*}[t]
      \centering
      \includegraphics[width=\linewidth]{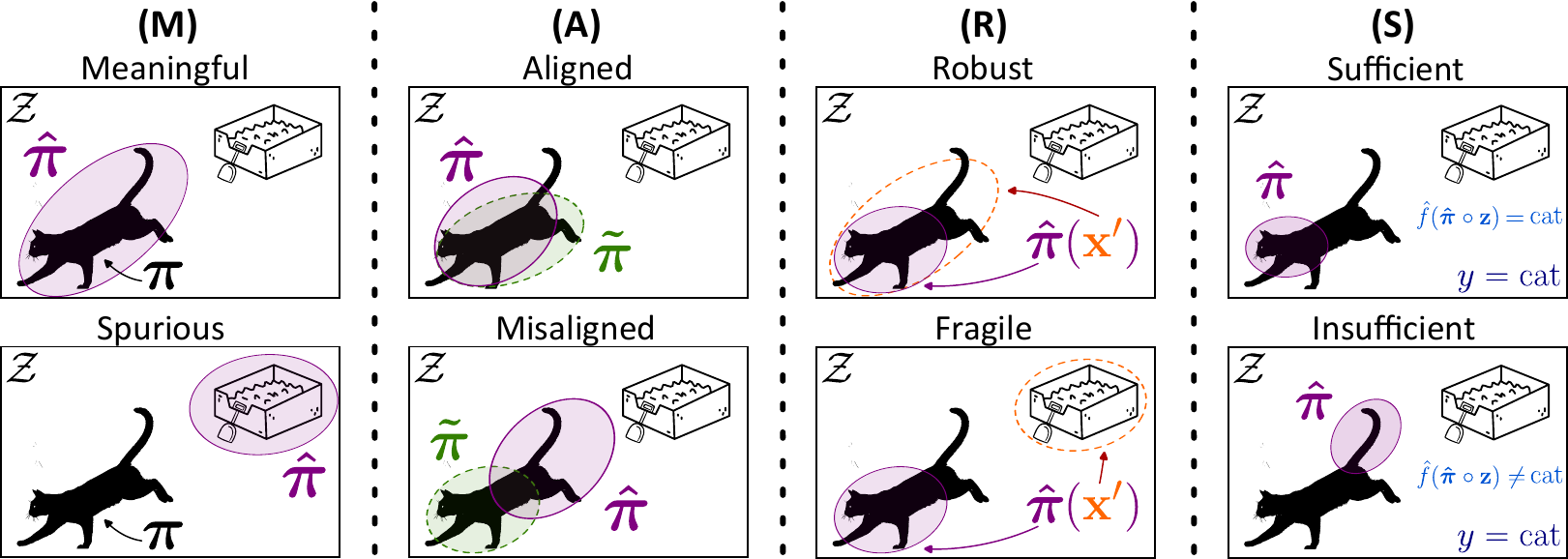}
      \caption{MARS criteria in a toy cat classification. Virtually, the image is a point in the space $\Z$ and is associated with a class prediction $\hat y$. Three types of attribution are represented: post-hoc estimate $\pibh$, intrinsic $\pib$, and ground-truth $\pibs$. The notation $(\x)$ is omitted for clarity, except in the (R)obustness column where $\x'\subseteq\x$ represents the subset of input features used to construct the explanation. For this criterion to be met, the signal $\x'$ used by the model to form the intrinsic attribution $\piboldhat(\x')$ must be highlighted by the latter. In the (M)eaningfulness, (R)obustness, and (S)ufficiency columns, $\piboldhat$ can be substituted with $\tilde\pibold$ as they apply to both intrinsic and post-hoc attributions. Sufficiency definition is tied to the predicted class $\tilde y$ as it must be retrievable from the explanation.} 
      \label{fig:mars}
    \end{figure*}

    \ptitle{Evaluating the faithfulness of explanations}
    Alignment alone cannot ensure that explanations capture meaningful signals in the data. A pseudo-linear model could, in principle, assign random-looking importance scores to features without compromising explanatory alignment and predictive accuracy. The challenge, especially when working with non-tabular data like images, is to design models whose intrinsic explanations are not only aligned but also faithful across other important dimensions. We introduce the \hlt{MARS} framework to evaluate explanatory faithfulness through four criteria (Fig. \ref{fig:mars}). We deem an explanation

    \begin{itemize}[itemsep=-2pt, topsep=3pt, labelwidth=1.6em, labelsep=0.5em, labelindent=.3cm, leftmargin=!]
        \renewcommand\makelabel[1]{\makebox[\labelwidth][c]{#1}}
        \item[\textbf{M}] \textbf{Meaningful} — if it detects relevant signals
        \item[\textbf{A}] \textbf{Aligned} — if it directly underlies the prediction
        \item[\textbf{R}] \textbf{Robust} — if it is insensitive to context
        \item[\textbf{S}] \textbf{Sufficient} — if it suffices to recover the prediction
    \end{itemize}

    To illustrate these criteria, consider classifying an image picturing both a cat and a litter box; the class to predict is \enquote{cat.} We denote the model's intrinsic, aligned attributions by $\pib$, post-hoc attributions by $\pibh$, and ground-truth, optimal attributions by $\pibs$. We assume that explanations are formed in the input space, i.e., $\Z\equiv\X$.

    {Meaningfulness} (Fig. \ref{fig:mars}, M) — An explanation is meaningful if it accurately captures what is relevant to the prediction. Given ground-truth attributions $\pibs$ (e.g., expert annotations highlighting the cat), meaningfulness requires $\pib\approx\pibs$ or $\pibh\approx\pibs$. In our example, a meaningful explanation should detect the pixels shaping the cat while ignoring spurious context like the litter box. 
    
    One can measure meaningfulness through accuracy statistics that compare $\pibh$ or $\pib$ to $\pibs$, when available. In the Methods section we define the \hlt{True Detection Rate} (TDR) — how much relevant signal is captured — and the \hlt{True Abstraction Rate} (TAR) — how much irrelevant signal is filtered out. A \hlt{detection score} is defined as the product of the two.

    {Alignment} (Fig. \ref{fig:mars}, A) — An explanation is aligned if it unambiguously reflects the construction of the prediction. For model-agnostic methods, we can only estimate $\pibh$ and hope $\pib$ indeed exists and is well approximated. Pseudo-linear models, in contrast, can produce aligned explanations $\pib$ by design. 

    {Robustness} (Fig. \ref{fig:mars}, R) — An explanation is robust if it does not depend heavily on contextual signal. If the explanation $\pib$ (or $\pibh$) relies on contextual signals like the litter box, removing those cues from input data can lead to altered explanations. A robust explanation is constructed from relevant signal alone, remaining stable when context is removed. 

    {Sufficiency} (Fig. \ref{fig:mars}, S) — An explanation is sufficient if it contains enough information to reconstruct the prediction \citep{SUFFICIENT1,SUFFICIENT2}. When we filter the input data using the attributions — keeping only features deemed important — we obtain the recursive input $\pib\circ\z$ (element-wise multiplication). Can we recover the original prediction by performing a recursive prediction $\hat f(\pib\circ\z)$? In our example, detecting the cat's tail alone may be insufficient to recover the class \enquote{cat.}, whereas the cat's head may suffice.

    We leverage recursive predictions to evaluate sufficiency and robustness. In light of its definition, a sufficient explanation yields a recursively stable prediction. On the other hand, a robust explanation should not rely on context, implying that the recursive explanation $\piboldhat(\pib\circ\z)$ should match the initial explanation $\pib$. Since PiNets construct predictions directly from explanations, a lack of recursive stability should manifest as shifts in predictive accuracy between initial and recursive predictions. We can therefore use the \hlt{recursive accuracy shift} as an indicator of sufficiency and robustness. 

    More information on the quantification of these criteria is provided in Methods.

    These criteria are rather complementary and explanations could score high on some dimensions while low on others. A meaningful explanation can be insufficient if misaligned; for example, a post-hoc explanation may highlight the cat while the model actually relied on the litter box. Conversely, an aligned-yet-spurious explanation might suffice for prediction reconstruction although it fails to capture semantically relevant signal. Finally, because what is highlighted by attributions $\pib$ and how they depend on features $\x$ need not be the same, meaningful explanations may lack robustness to context. 
    %These examples shed light on the divide between (i) what the data really tells, (ii) what patterns the model learns to exploit, and (iii) whether we have access to the model's explanations. 

    \begin{figure*}[!t]
      \centering
      \includegraphics[width=\linewidth]{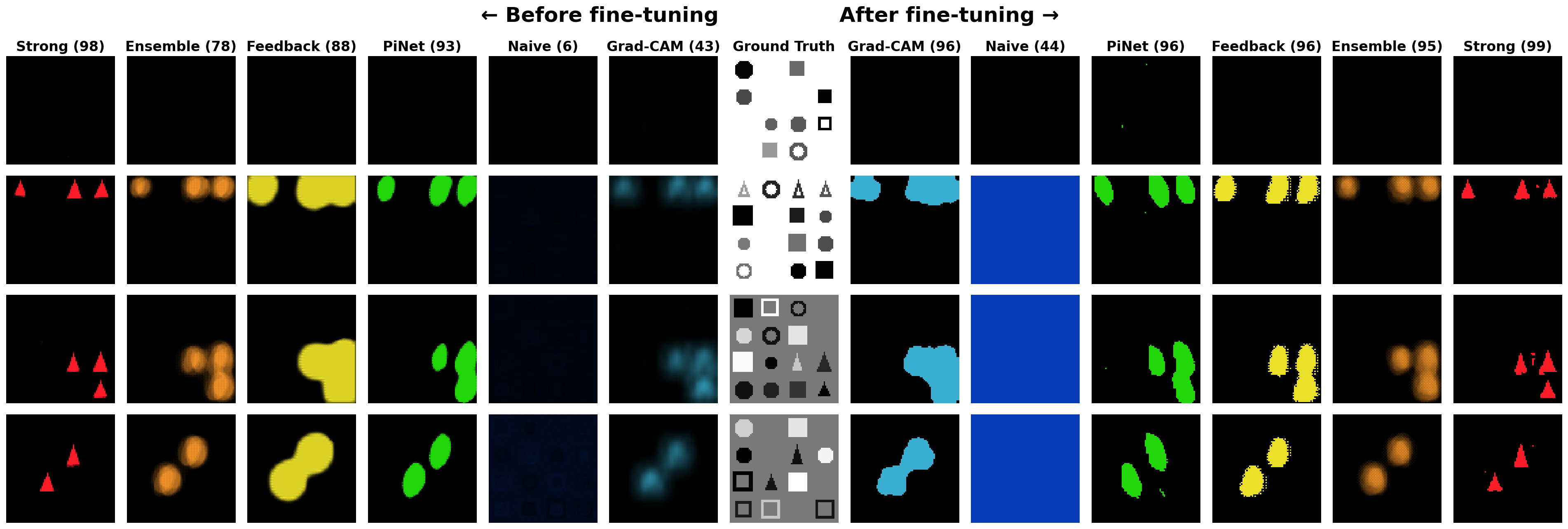}
      \caption{Test set attribution examples in ToyShapes for the top explainer of each model group, selected based on average meaningfulness (reported in parentheses). The soft second look variant is omitted as it produces results very similar to the default variant.} 
      \label{fig:toyshapes_detection}
    \end{figure*}
    
    \begin{figure*}[!b]
      \centering
      \includegraphics[width=0.7\linewidth]{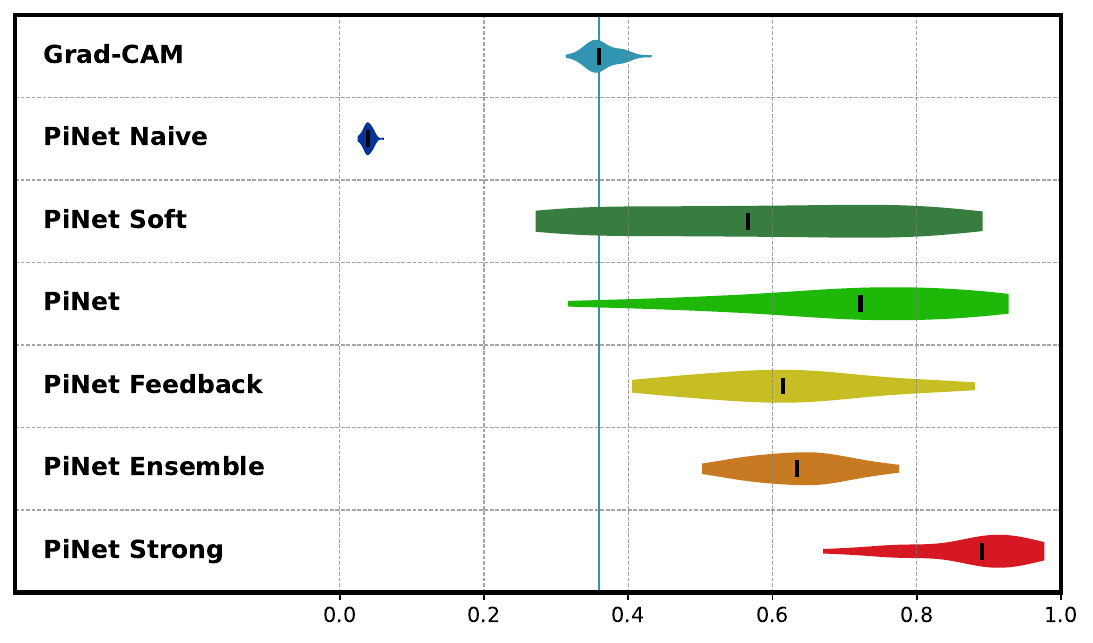}
      \caption{Meaningfulness before fine-tuning in ToyShapes depicted by violin plots. Marks inside each violin represent medians. Blue vertical lines extend the medians of the baseline (\fdot[Ccam]{1.5} Grad-CAMs). Aside from the {\fdot[Cnaive]{1.5}} PiNet Naive variant (inadequate decoder), PiNets consistently outperform Grad-CAMs before fine-tuning.} 
      \label{fig:toyshapes_perf}
    \end{figure*}
    
    \begin{figure*}[!t]
      \centering
      \includegraphics[width=\linewidth]{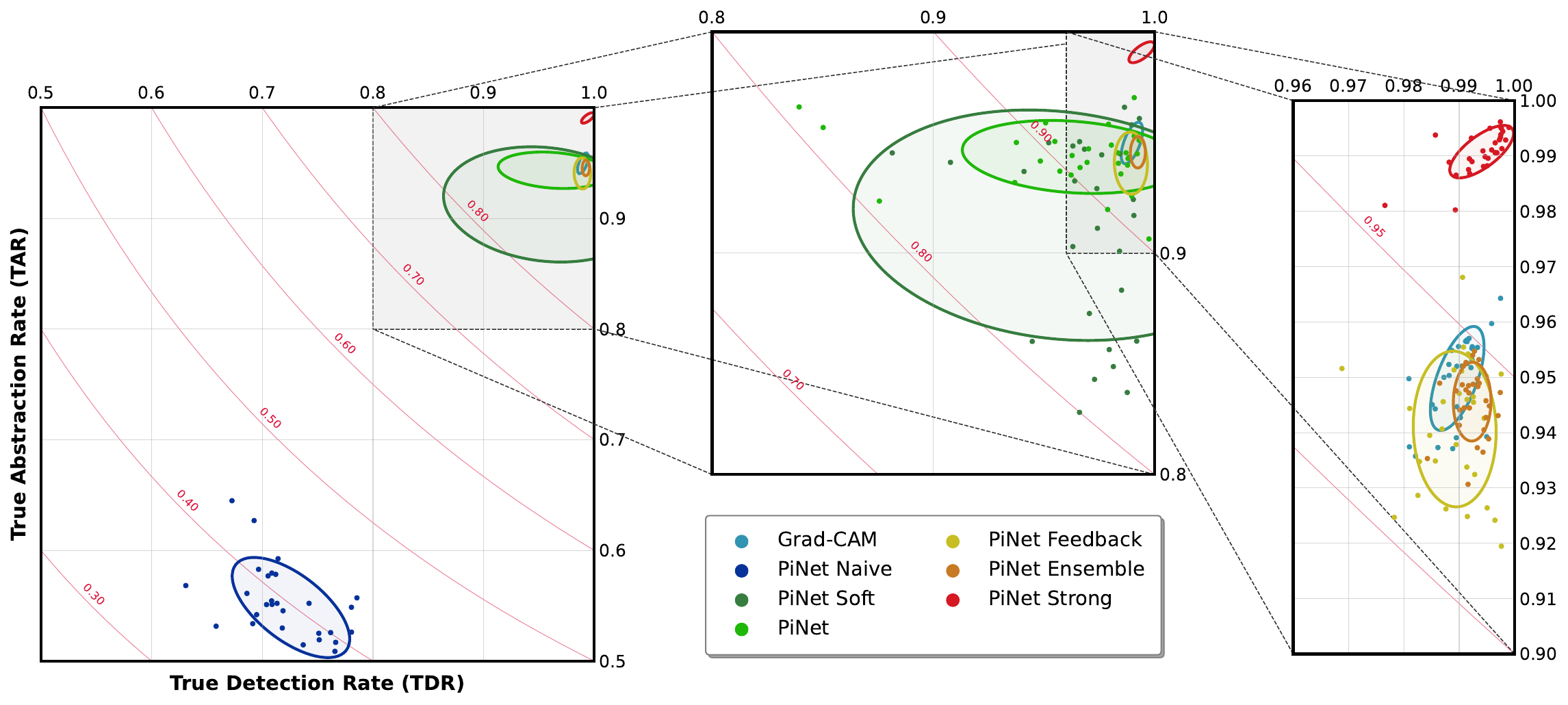}
      \caption{Meaningfulness of fine-tuned (thresholded) attributions in ToyShapes. Red contours show iso-levels of the detection score (eq. \eqref{eq:score}). Dots represent trained models color-coded by variant, and are shown together with 95\% Gaussian confidence ellipses. Ellipses inside the gray insets (left and middle panels) are shown without dots for visual clarity; the same insets are magnified in the panel on their right. When fine-tuning is possible PiNets perform on par with the baseline (\fdot[Ccam]{1.5} Grad-CAMs) only when enhanced with techniques like \fdot[Cfeedback]{1.5} recursive stabilization or \fdot[Censemble]{1.5} ensembling. \fdot[Cstrong]{1.5} Strong supervision, if feasible, yields the best results.}
      \label{fig:toyshapes_perf_optim}
    \end{figure*}

    \hfill\\
    \ptitle{PiNets}
    We present \hlt{Pointwise-interpretable Networks} (PiNets), a neural network architecture designed to achieve explanatory alignment through pseudo-linearity. Unlike standard networks that learn complex internal features $\h(\x)$ and combine them linearly with fixed coefficients, PiNets learn coefficient functions $\pib$ and combine them linearly with user-specified interpretable features $\z$.

    A PiNet comprises four components: an \hlt{encoder} extracting internal features, or encodings, $\h(\x)$ from inputs $\x$; a \hlt{decoder} transforming encodings $\h(\x)$ into coefficients $\pib$; a \hlt{second look} operation — the key mechanism for immediate precedence — that applies these coefficients to interpretable features $\z$ via element-wise multiplication; and a linear \hlt{aggregator} producing predictions (Fig. \ref{fig:pinet}). For a prediction problem involving the construction of multiple outputs, each output $\hat y_k$ is constructed as:
    \begin{equation}
        \hat y_k = \hat a_k + \hat b_k \sum\nolimits_* \pibk \circ \z
    \end{equation}
    where $\hat a_k$ and $\hat b_k$ are optional shift and scale parameters. Hence, each output $\hat y_k$ is assigned its own set of coefficients $\pibk$. In classification problems, where $\{\hat y_k\}_k$ are logits, one could call \enquote{explanations} those coefficients associated with the predicted classes and discard the others.

    \vspace{.2cm}
    \begin{figcol}
      \centering
      \includegraphics[width=.8\textwidth]{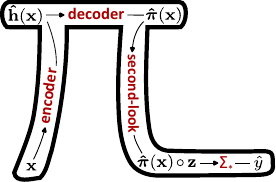}
      \captionof{figure}{PiNet architecture}
      \label{fig:pinet}
    \end{figcol} 
    
    When $\z$ is fully interpretable, the coefficients $\pib$ constitute aligned explanations: they directly and transparently underlie predictions through a linear operation.

    Critical design choices include the interpretable feature space $\Z$ (e.g., should it coincide with $\X$?) and the encoder-decoder architecture that must enable both predictive accuracy and explanatory faithfulness. Moreover, while the second look mechanism (explicit multiplication $\pib \circ \z$) is central to pseudo-linearity, it can be relaxed through a soft variant when feature values are uninformative for prediction construction. In such cases, replacing $\z$ with ones yields $\hat y_k = \hat a_k + \hat b_k \sum_* \pibk$, where coefficients are still constrained to lie in $\Z$. 

    In what follows, we let $\X$ and $\Z$ coincide and focus on the core architectural principles, deferring discussion of alternative feature spaces to the Discussion. We explore how the encoder-decoder design and several training techniques shape explanatory faithfulness, using a synthetic image dataset (ToyShapes) that provides controlled access to ground-truth attributions $\pibs$ for rigorous evaluation of meaningfulness. 

    It is worth stressing that PiNets are trained just like traditional prediction networks; i.e., to minimize a loss function reflecting predictive accuracy (e.g., cross-entropy).

    The construction of ToyShapes, the processing of attributions, and the experimental settings for evaluating PiNets against the baseline post-hoc method (Grad-CAMs, \citep{GRADCAM}) are provided in Methods. Several variants of PiNets are considered and consistently color-coded across visualizations and the text (see the list in Methods and a minimal version of it in the legend of Figure \ref{fig:toyshapes_perf_optim}).

    \begin{figure*}[!b]
      \centering
      \includegraphics[width=.9\linewidth]{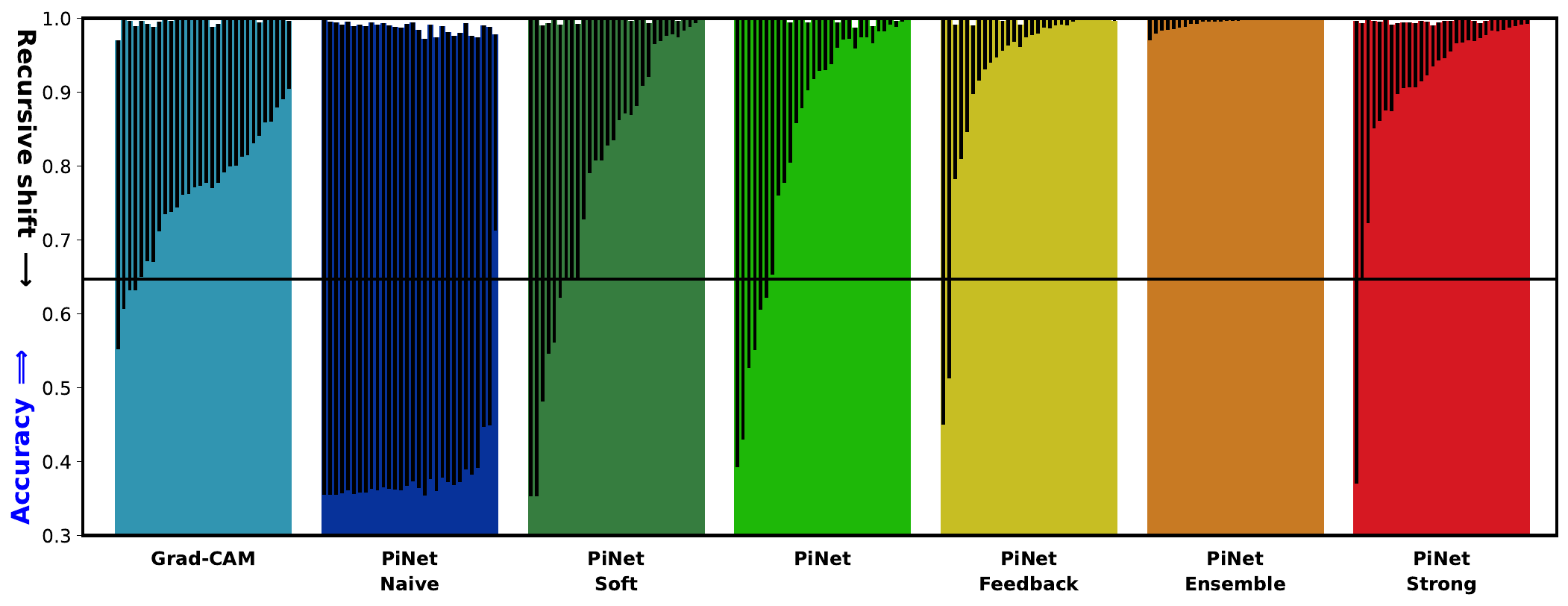}
      \caption{Recursive accuracy shift in ToyShapes. 
      The colored vertical bars in each panel represent test predictive accuracy (vertical axis starts at 0.3). The black sticks represent recursive accuracy shifts; they start at the accuracy level on top, and extend downwards to reach the accuracy levels achieved after recursion. The horizontal black line shows the naive predictive accuracy level achieved by always predicting the dominant class (presence of triangles). Recursive accuracy below this level indicates a tendency to predict the absence of triangles from the recursive input, a sign of gross insufficiency of explanations. PiNets improve upon the baseline (\fdot[Ccam]{1.5} Grad-CAMs), with the most striking improvements observed for the \fdot[Censemble]{1.5} PiNet Ensemble and  \fdot[Cfeedback]{1.5} PiNet Feedback variants.}
      \label{fig:toyshapes_recursive}
    \end{figure*}
    
    \hfill\\
    \ptitle{The key role of the decoder}
    Having established PiNets' anatomy, we turn to a critical question: how does architectural design shape the model's capacity to produce faithful explanations? 

    The encoder-decoder architecture embodies a fundamental division of labor: the encoder acquires statistical intelligence from inputs $\x$ and the decoder imposes structure (based on $\Z$) on the way this intelligence is communicated and used to construct predictions.

    This distinction proves critical. Consider the {\fdot[Cnaive]{1.5}} PiNet Naive variant, whose decoder is a simple layer of perceptrons inadequate for spatial semantics. Despite achieving high classification accuracy on ToyShapes, its explanations are spurious: the detection score (reflecting meaningfulness) hovers near-chance level (Fig. \ref{fig:toyshapes_perf}, \ref{fig:toyshapes_perf_optim}, and \ref{fig:toyshapes_detection}). 

    This failure exposes a critical insight: predictive accuracy and explanatory meaningfulness are not automatically coupled in pseudo-linear networks. Without adequate architectural constraints, the model finds computational shortcuts — carrying early encoder's predictions throughout the decoder, and building coefficients $\pib$ that technically underlie predictions but carry no semantic content (Fig. \ref{fig:toyshapes_detection}). 

    All other PiNet variants employ a symmetric encoder-decoder architecture with transposed convolutions and achieve substantially higher detection scores (Fig. \ref{fig:toyshapes_perf} and \ref{fig:toyshapes_perf_optim}). Visual inspection confirms spatially-coherent attribution that accurately detect target objects (Fig. \ref{fig:toyshapes_detection}).

    The decoder constrains the model to organize its learned intelligence into spatially-coherent patterns. This constraint makes meaningful explanations not merely possible but necessary — the model cannot anymore achieve high accuracy without structuring $\pib$ into meaningful detections. This inverts the typical explainable AI paradigm: rather than extracting explanations from black-box models through post-hoc methods, we design models wherein explanations and predictions are architecturally intertwined.

    \begin{figure*}[!b]
      \centering
      \includegraphics[width=.9\linewidth]{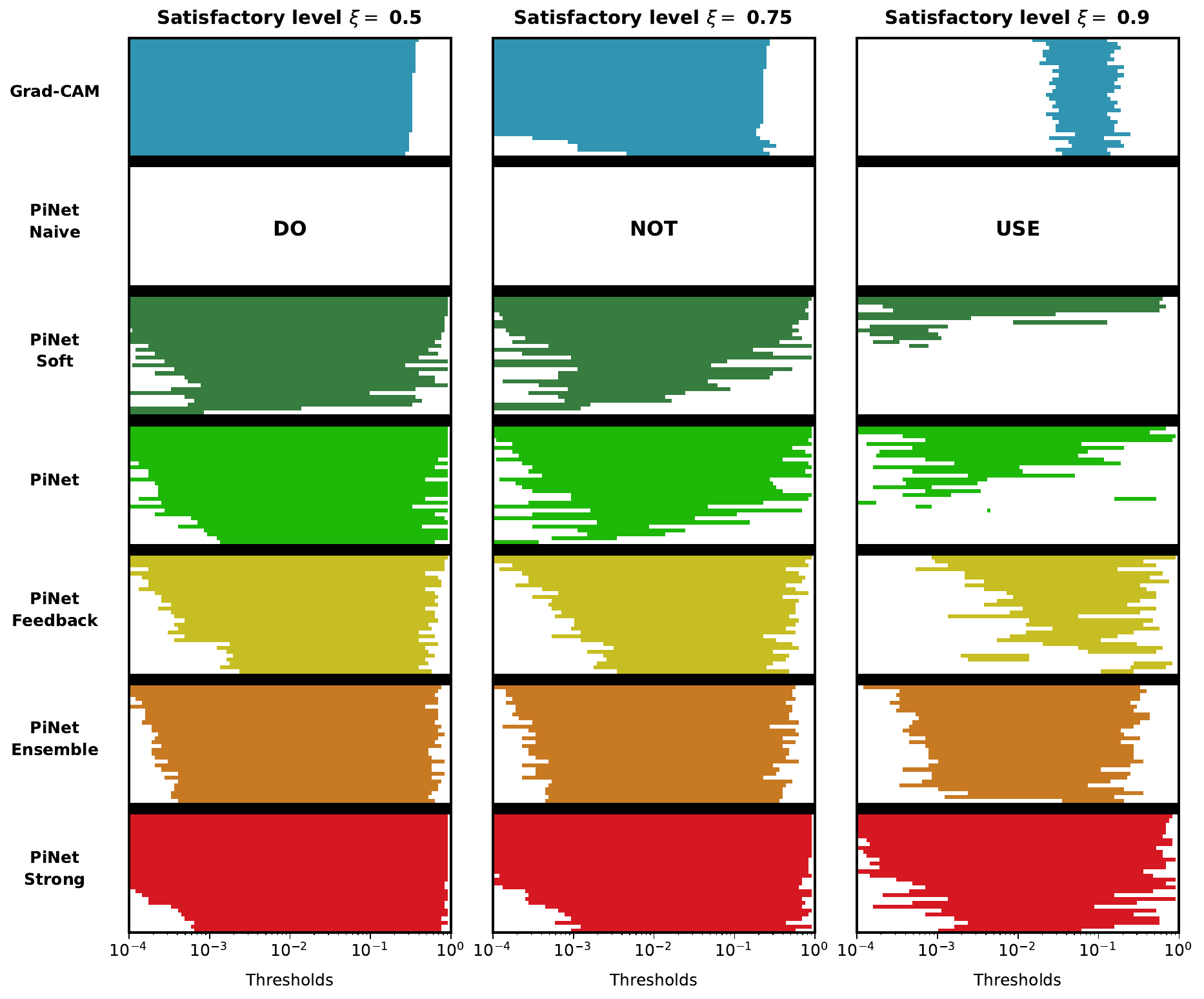}
      \caption{Ease of fine-tuning meaningfulness in ToyShapes. Bars represent the ranges of thresholds satisfying a detection score of at least $\xi$ (in the column title). Thresholds reported on the x-axis are log-transformed and range from $10^{-4}$ to $1$. Within each model group, results are sorted by the breadth of the range, for clarity. The {\fdot[Cnaive]{1.5}} PiNet Naive variant is flagged \enquote{do not use} as it yields spurious explanations for any threshold $t\in[0,1]$. Enhanced PiNets, especially with \fdot[Censemble]{1.5} ensembling and \fdot[Cstrong]{1.5} strong supervision, appear easier to fine-tune and thus to yield quality explanations when ground-truth attributions are not available for meaningfulness quantification.}
      \label{fig:toyshapes_range}
    \end{figure*}

    \hfill\\
    \ptitle{How to look twice?}
    The second look mechanism ($\pib \circ \z$) ensures immediate precedence by explicitly applying coefficients to features. However, when feature values are uninformative for prediction construction (e.g., in images where pixel intensities vary independently of class labels), we can relax this to the soft variant: replacing $\z$ with ones such that $y=\sum_* \pib$. Importantly, coefficients still lie in $\Z$, preserving architectural constraints on their spatial structure.

    In ToyShapes, the default {\fdot[Cpinet]{1.5}} PiNet variant employs a hard second look and further improves results stability compared to the {\fdot[Csoft]{1.5}} PiNet Soft variant (Fig. \ref{fig:toyshapes_perf} and \ref{fig:toyshapes_perf_optim}). The hard variant provides additional stabilization when values can serve as anchors but is not strictly necessary. However, were we forming explanations in a tabular feature space, the hard second look would be essential to ensure that coefficients remain consistently paired with the same features.

    \hfill\\
    \ptitle{Recursive stabilization}
    Predictions and explanations should remain consistent when the model examines only the signal it deems relevant based on the attributions. In PiNets, we can build this in through a feedback mechanism that penalizes dissimilarity between initial coefficients $\pib$ and recursive coefficients $\piboldhat(\pib\circ\z)$ — those generated when the model predicts from filtered inputs $\pib\circ\z$ (Fig. \ref{fig:pinet_feedback} in Methods).

    By constraining explanations to remain stable under recursion, this mechanism directly seeks robustness and sufficiency. Robustness improves because context gets filtered out from the recursive input ($\pib\circ\z$) and thus cannot be used to construct the recursive explanation. Sufficiency improves because prediction stability is implied by explanation stability in PiNets.

    Results (Fig. \ref{fig:toyshapes_recursive}) indeed indicate that the {\fdot[Cfeedback]{1.5}} PiNet Feedback variant exhibits substantially smaller and less variable accuracy shifts under recursive prediction compared to {\fdot[Cpinet]{1.5}} default PiNets.

    \begin{figure*}[!b]
      \centering
      \includegraphics[width=.9\linewidth]{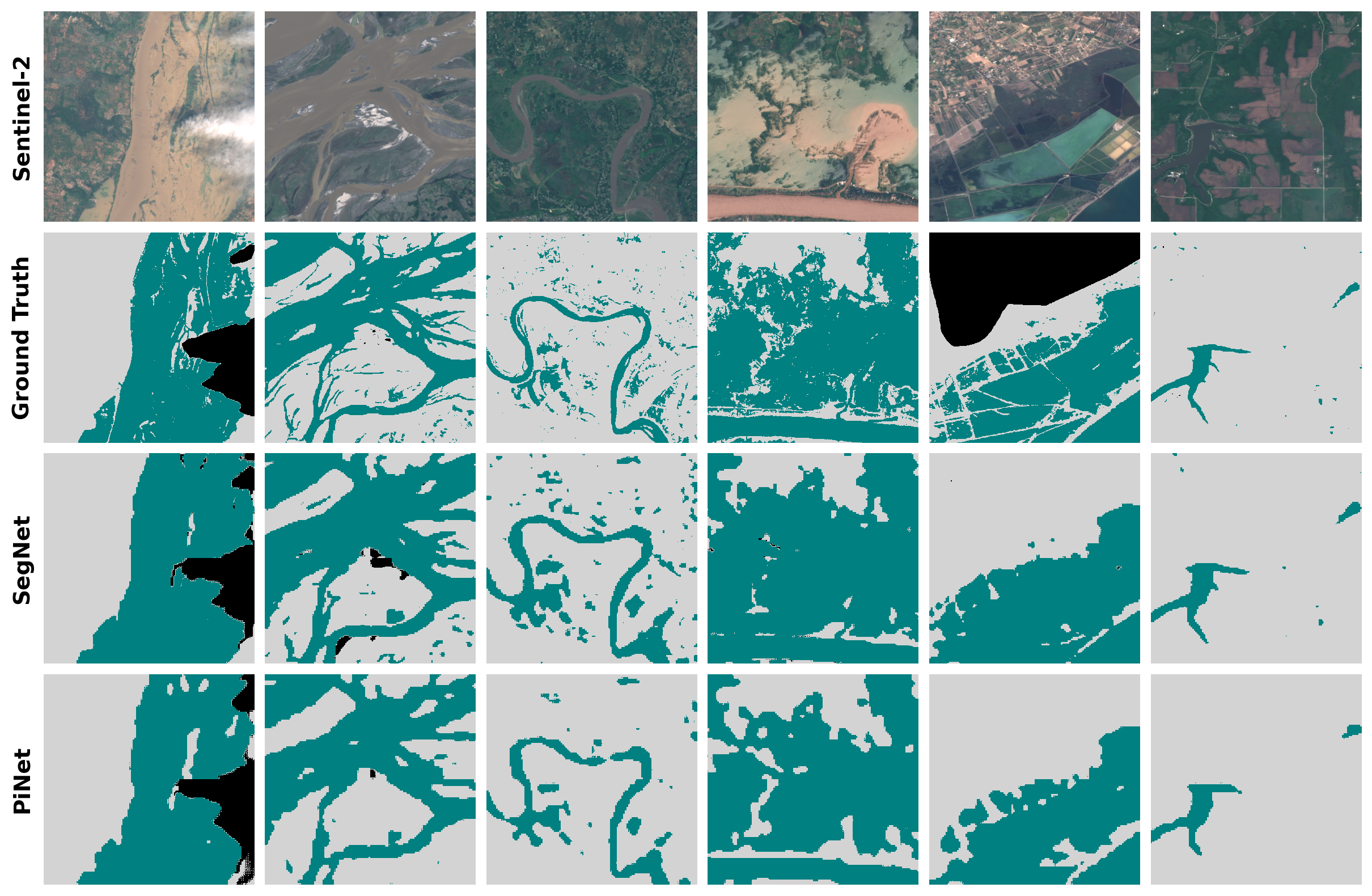}
      \caption{Test segmentation maps in Sen1Floods11. From top to bottom: Sentinel-2 images (RGB bands), hand-annotated maps (proxy for ground-truth), SegNet segmentation maps, and PiNet attribution maps. Classes $k=\set{-1,0,1}$ are assigned, respectively, the colors black (no data/not valid), gray (no water) and teal (water).
      }
      \label{fig:floods}
    \end{figure*}

    \hfill\\
    \ptitle{Ensembling} 
    Averaging predictions across multiple models typically improves accuracy by diluting or evening out individual errors. But ensembles often sacrifice explainability — blending readable components into an opaque aggregate (like random forests \citep{FOREST}). In contrast, an additive ensemble of PiNets collapses into a single pseudo-linear network, preserving readability and alignment (see Methods). 

    Beyond its effect on predictive accuracy, ensembling also stabilizes explanations by smoothing out model-specific errors. The {\fdot[Censemble]{1.5}} PiNet Ensemble variant shows improved stability in both predictive accuracy (Fig. \ref{fig:toyshapes_recursive}) and explanatory faithfulness (Fig. \ref{fig:toyshapes_perf} and \ref{fig:toyshapes_perf_optim}). Most strikingly, ensembles display significantly better recursive stability (Fig. \ref{fig:toyshapes_recursive}). Ensembling PiNets is therefore a promising approach to improving explanatory meaningfulness, sufficiency, and robustness in settings where alignment can be guaranteed.

    \hfill\\
    \ptitle{Strong supervision}
    When ground-truth attributions $\pibs$ are available, PiNets' training can be directly supervised on those. The training loss is then augmented with an attribution loss (see Methods) measuring the meaningfulness (accuracy) of model explanations $\pib$ with respect to $\pibs$.

    This approach requires careful consideration: the quality and representativeness of ground-truth annotations directly determines the quality of learned explanations. Biased or incomplete annotations will produce correspondingly flawed explanations. Yet when annotations are trustworthy, strong supervision yields dramatic improvements. The {\fdot[Cstrong]{1.5}} PiNet Strong variant indeed achieves near-perfect meaningfulness in ToyShapes (Fig. \ref{fig:toyshapes_perf}, \ref{fig:toyshapes_perf_optim}, and \ref{fig:toyshapes_detection}).

    Even limited supervision can be effective. In ToyShapes experiments, exposing PiNets to just 25 ground-truth maps (3\% of training examples) substantially improved explanatory faithfulness without compromising predictive accuracy.

    \hfill\\
    \ptitle{Attribution fine-tuning} The detection score (representing explanatory meaningfulness) is tied to the sharpness and confidence of attributions. Fine-tuned attributions, whose performance is reported in Fig. \ref{fig:toyshapes_perf_optim}, stem from the fine-tuning of a threshold applied to the attributions $\pib$ (see Methods). 

    A practical question arises: without ground-truth attributions $\pibs$ — required for the quantitative assessment of meaningfulness — how difficult is it to fine-tune PiNet explanations and achieve satisfactory detection quality? We addressed this by measuring the range of thresholds that yield detection scores above an arbitrary value. A wider range indicates easier fine-tuning: one should achieve satisfactory explanations faster without extensive threshold search. Because in practice meaningfulness assessment can require tedious, qualitative (e.g. visual) inspection, it is essential to obtain large ranges of satisfactory thresholds.
    
    For moderate quality targets, both PiNets and Grad-CAMs provide wide threshold ranges (Fig. \ref{fig:toyshapes_range}). However, a striking pattern emerges when targeting high-quality explanations (detection score $\geq$ 0.9). Grad-CAMs show much narrower ranges than PiNets, and this pattern get more consistent in the enhanced PiNet variants.
    
    Add to this that without thresholding, i.e., before fine-tuning, all PiNet variants (except the naive-decoder variant) perform considerably better that Grad-CAMs. Overall, converging on more meaningful attributions is less costly with PiNets.    

    % \hfill\\
    \columnbreak
    \ptitle{Computational cost}
    Across ToyShapes experiments we observe that PiNets require roughly twice as long to train compared to baseline models, a cost associated with the decoder module. However, once trained, the production of explanations is nearly instantaneous, as they are generated by the model just before predictions. Gradient-based methods require gradient computations which substantially increase the time needed to produce attributions. Besides, model-agnostic approaches like SHAP and LIME are even more computationally intensive as they require computing a large amount of predictions to estimate local feature importance \citep{PERTURB2}. For this reason, PiNets could stand out in applications where trustworthy explanations are required in real-time or on a large scale.

    \hfill\\
    \ptitle{Precision in targets induces sharpness in explanations}
    We stressed that principled decoder architectures constrain PiNets toward meaningful explanations as a precondition for predictive accuracy. This raises a key question: could more informative prediction targets further sharpen this constraint? In classification, where only class presence matters, many decision boundaries achieve equivalent predictive accuracy levels — requiring only moderate sharpness in attributions. Would a regression task, implying precise quantitative predictions, compel PiNets to organize sharper explanations?

    We explore this question through a flood mapping exercise from satellite imagery using the Sen1Floods11 dataset \citep{FLOODS}. We train two models with identical encoder-decoder architectures: a SegNet (baseline) supervised with pixel-level segmentation maps, and a PiNet supervised only with image-level surface area targets — the count of flooded versus non-flooded pixels. This design isolates the effect of target precision on attribution sharpness while maintaining architectural parity. Indeed, the PiNet is really a SegNet with a different aggregator and loss function, and all training and modeling settings are identical (more about the data and settings in Methods). 

    Despite receiving coarser supervision, the PiNet produces competitive segmentations. While the SegNet achieves sharper water detection — expected given its access to finer supervision — the attributions of the PiNet still provide marked spatial coherence (Table \ref{tab:floods} and Fig. \ref{fig:floods}). 

    \needspace{1cm}
    Importantly, PiNet attributions show greater sharpness than in ToyShapes classifications. This confirms our hypothesis: more informative prediction targets constrain PiNets to organize explanations with greater sharpness. The regression objective — predicting pixel counts rather than object presence — mechanically demands that attributions capture finer spatial detail to construct accurate predictions.

    These results demonstrate PiNets' potential for segmentation under weak supervision. When pixel-level annotations are scarce but descriptive image-level targets are available — like surface areas or object counts — PiNets may offer a principled path to affordable semantic segmentation. 

    {\renewcommand{\arraystretch}{1.5} 
    \small
    \setlength{\arrayrulewidth}{.1em}
    \begin{center}
        \centering \vspace{.2cm}
        \begin{tabular}{
            >{\columncolor[HTML]{EFEFEF}}l |
            c c c |
            c c c}
            % Header row with dataset labels
            \rowcolor[HTML]{EFEFEF}
            & \multicolumn{3}{c|}{\textbf{Water}} 
            & \multicolumn{2}{c}{\textbf{No water}} \\
            % \cline{2-6}
            % Metric column headers
            \rowcolor[HTML]{EFEFEF}
            & \textbf{MAE} $\thickdownarrow$ 
            & \textbf{IoU} $\thickuparrow$ 
            & \textbf{TDR} $\thickuparrow$ 
            & \textbf{IoU} $\thickuparrow$ 
            & \textbf{TDR} $\thickuparrow$ \\
            \hline
            % Data rows
            \textbf{SegNet} & 2582 & 0.332 & 0.904 & 0.814 & 0.940 \\
            \textbf{PiNet}  & 1110 & 0.256 & 0.802 & 0.819 & 0.959 \\
            \Xhline{.5\arrayrulewidth}
            \textbf{Delta}  & \better{-57.0\%} & \worse{-22.9\%} & \worse{-11.3\%} & \better{+0.6\%} & \better{+2.0\%} \\
        \end{tabular}
        \captionsetup{width=.9\linewidth, font={small}}
        \captionof{table}{Performance in flood mapping. The level of improvement (\better{green}) or deterioration (\worse{red}) of the PiNet over the SegNet is reported in the Delta row. $\protect\thickuparrow$ and $\protect\thickdownarrow$ indicate the higher the better and the lower the better, respectively.}
        \label{tab:floods}
    \end{center}}

\end{multicols}

%%%%%%%%%%%%%%%%%%%%%%%%%%%%%%%%%%%%%%%%%%%%%%%%%%%%%%%%%%%%%%%%%%%%%%%
%%%%%%%%%%%%%%%%%%%%%%%%%%%%%%%%%%%%%%%%%%%%%%%%%%%%%%%%%%%%%%%%%%%%%%%
%%%%%%%%%%%%%%%%%%%%%%%%%%%%%%%%%%%%%%%%%%%%%%%%%%%%%%%%%%%%%%%%%%%%%%%

\needspace{3cm}
\section{Discussion}
\label{sec:sec3}

\begin{multicols}{2}
\setlength{\parskip}{0pt}

    The ability to explain how a prediction is made is not merely a desirable property of AI systems — it is a precondition for trusting them. Nonetheless, popular explanation methods either rationalize predictions rather than genuinely underlie them or, when they do underlie them, do so ambiguously. PiNets address this flaw at the architectural level. By shifting the black box from the features to the coefficients PiNets are linearly readable and guarantee explanatory alignment insofar as the linearly combined features are interpretable.

    The results across our experiments reveal a consistent and important lesson: in PiNets, the quality of explanations is not independent of predictive accuracy — it is architecturally coupled to it. Equipped with a naive decoder, PiNets find their way to accurate predictions while producing spurious explanations. This demonstrates that alignment does not guarantee meaningfulness. The encoder-decoder design closes this loophole by constraining the transformation from rich encodings $\h(\x)$ to coefficients $\pib$ to preserve spatial structure, making meaningful explanations a prerequisite for accurate predictions rather than a byproduct of them. 
    
    To further improve the faithfulness of explanations we introduced and validated a set of training techniques. Recursive stabilization directly targets robustness and sufficiency by enforcing consistency between initial and recursive explanations. Ensembling has a variance-reduction effect on both predictions and attributions, yields greater explanatory faithfulness, and does not break alignment. Finally, when ground-truth attributions are available, even in limited amount, strong supervision provides a powerful lever to directly optimize explanation quality. These enhancements substantially improve the meaningfulness, sufficiency, and robustness of PiNets' explanations. We also find that fine-tuning attributions for greater meaningfulness is easier in PiNets than in Grad-CAMs, suggesting that practitioners can more easily converge on high-quality explanations when little or no ground-truth attributions are available for quantitative evaluation.

    Finally, we showed that the precision of prediction targets shapes the sharpness of explanations. In a flood mapping exercise, PiNets organize sharp segmentations when regressing surface area. This suggests that PiNets are particularly well-suited for learning meaningful spatial attributions when descriptive image-level targets are available but pixel-level annotations are scarce.

    PiNets' dual capacity to form predictions and explanations positions them advantageously for practical applications. In segmentation tasks with limited annotations, PiNets could overcome the tedious multi-stage training procedures required for weak supervision \citep{WEAKLY1,WEAKLY2}. 

    The notion of robustness to context we introduced in the MARS evaluation framework effectively extends the classic problem of reliance on spurious signal in prediction \citep{SPURIOUS1,SPURIOUS2,SPURIOUS3} to explanation. It warrants deeper investigation. While distribution shifts and adversarial perturbations have been extensively studied for predictions \citep{DATASHIFT,DOMAIN,ADVERSARIAL,OOD}, explanatory robustness remains underexplored. Future work could examine whether techniques like adversarial training \citep{ADVERSARIALNET} or data augmentation \citep{MIXUP,CUTMIX} can enhance the out-of-distribution stability of both predictions and explanations.

    The flexibility of PiNets extends naturally to diverse data modalities through the separation of input space $\X$ and explanation space $\Z$. This architectural choice enables hybrid modeling strategies where raw, potentially multimodal data — images, genomic sequences, sensor streams — fuel expressive neural encoders, while explanations are constrained to interpretable representations aligned with domain knowledge. For sequential data, raw sequences in $\X$ could drive deep encoders while explanations in $\Z$ leverage spectrograms for audio or functional motifs for genomic data. In graph-structured problems, graph neural networks could extract rich topological features from raw graphs in $\X$ while explanations in $\Z$ reference interpretable graph properties. Interestingly, explanations can be formed over curated tabular feature spaces encoding domain expertise, even when predictions are constructed from complex non-tabular inputs.

    This architectural flexibility admits two equivalent yet conceptually distinct interpretations. The first, adopted throughout this paper, frames PiNets as neural networks with linear structure — advancing explainable AI by reconciling deep learning's predictive power with transparent decision-making. The second interprets them as linear models augmented with coefficient functions capable of integrating complex data previously inaccessible to classical statistical frameworks. While mathematically equivalent, these perspectives carry different implications: the former promises more trustworthy AI systems, while the latter opens new avenues for scientific discovery. By constraining explanations to interpretable feature spaces that encode domain expertise, PiNets could unveil insights that neither traditional linear models nor opaque neural networks alone could reveal — effectively bridging modern machine learning with the interpretive rigor of classical statistics.

\end{multicols}

%%%%%%%%%%%%%%%%%%%%%%%%%%%%%%%%%%%%%%%%%%%%%%%%%%%%%%%%%%%%%%%%%%%%%%%
%%%%%%%%%%%%%%%%%%%%%%%%%%%%%%%%%%%%%%%%%%%%%%%%%%%%%%%%%%%%%%%%%%%%%%%
%%%%%%%%%%%%%%%%%%%%%%%%%%%%%%%%%%%%%%%%%%%%%%%%%%%%%%%%%%%%%%%%%%%%%%%

\needspace{3cm}
\section{Methods}
\label{sec:sec4}

\begin{multicols}{2}
\setlength{\parskip}{0pt}
\captionsetup{width=.9\linewidth, font={small}}

    \ptitle{Definitions} In the following we formalize alignment, readability, and pseudo-linearity. 

    \begin{formal}
      \textbf{Definition 1. Aligned explanation} \\[2pt]
      Explanatory alignment at point $(\x\in\X,\ \z\in\Z,\ \hat{y} = \hat f(\x))$ is possible if $\z$ is fully interpretable and the model $\hat f$ embeds both the feature attributions $\pib\in\Z$ and a simple function $g:\Z^2\to\Y$ such that $\hat{y} = g(\pib, \z)$. Then, the explanations represented by features attributions $\pibh=\pib$ are perfectly aligned.
    \end{formal}

    Intrinscness and immediate precedence are implied by this definition even if attributions are estimated post hoc since one cannot approximate or recover an explanation that does not exist in the first place. It follows that, whereas the meaningfulness of feature attributions can be a side effect of learning accurate predictions \citep{ACCANDXAI,ACCANDXAI2}, alignment cannot arise from predictive performance. Alignment requires that we work with models that produce attributions intrinsically, which points to model readability.  
    
    \begin{formal}
      \textbf{Definition 2. Readable model} \\[2pt]
      A model $\hat f:\X\to\Y$ producing predictions $\hat{y} = \hat f(\x)$ is deemed readable if it can be rewritten in the form $\hat{y} = g(\pib, \z)$, where $\z\in\Z$ is fully interpretable and $g:\Z^2\to\Y$ is a simple aggregation function. 
    \end{formal}

    Together, the interpretability of features $\z$ and the immediate precedence of explanations over predictions enforced by the readable function $g:\Z^2\to\Y$, guarantee explanatory alignment.

    Let us consider some classic examples. (Generalized) linear models produce predictions by combining input features $\x$ through a linear function $\hat f(\x)=\sum\nolimits_*\piboldhat\circ\x$. It naturally follows that $g(\piboldhat,\x) = f(\x)$. In spite of the inherent readability of such models, alignment can be hindered by a high-dimensional feature space \citep{MYTHOS}, oftentimes necessary to achieve reasonable levels of predictive accuracy in linear predictions. 
    
    In contrast, decision trees typically partition a lower-dimensional feature space since they can learn non-linearities and interactions internally \citep{CART}. However, it is their functional form (the tree) that grows in complexity to better fit the data. Consequently, the readability of tree-based models — whether deep individual trees or ensembles such as random forests \citep{FOREST} — is sometimes compromised by their complex structure.

    Interestingly, judicious integrations of linear and tree-based models can improve readability while offering reasonable predictive accuracy. For example, the RuleFit algorithm \citep{RULEFIT} estimates a linear model (simple $g$) over decision rules learned by tree-based models (interpretable $\z$). 

    \begin{formal}
      \textbf{Definition 3. Pseudo-linear model} \\[2pt]
      Let $\sum_*$ and $\circ$ denote, respectively, element-wise summation and multiplication, and let us consider the two observable feature sets $\x\in\X$ and $\z\in\Z$ (we can have $\X\equiv\Z$). A pseudo-linear model takes the generic form:
      \begin{equation}
        \hat{y} = \hat a + \sum\nolimits_* \pib \circ \z
        \label{eq:plm}
      \end{equation}
      \vspace{-0.5cm}
    \end{formal}

    Such models belong to the class of varying-coefficient models (VCMs \citep{VCM,VCM2,SURVEY3}) and we call {pseudo-linear models} their linear subclass wherein $\piboldhat:\X\to\Z$ is a varying-coefficient function mapping the input features $\x$ into coefficients that lie in the user-defined feature space $\Z$, and where both are combined linearly to construct $y$.

    Pseudo-linear models are linearly readable by design as long as $\Z$ is fully interpretable. By constructing linear models instance-wise, any input $\x\in\X$ gives rise to a set of varying coefficients $\pib$ which are then used to build the prediction $\hat{y} = g(\pib, \z)$ through a linear aggregator $g$. 

    The mechanism enabling readability is the {second look}. Coefficients $\pib\in\Z$ are applied to features $\z\in\Z$, an operation explicitly asking the model to consider the data again (through $\Z$) after extracting information from it (through $\X$). The second look can be seen as a mechanistic operation enabling immediate precedence; that is, it must be explicitly performed in the last layer of the network. For example, while B-cos networks \citep{BCOS} can be rewritten in a pseudo-linear form (at least when $\Z\equiv\X$) they do not rely on a second look in the last layer and do not satisfy immediate precedence. The second look really occurs in the first layer and the pseudo-linear form appears once we collapse every subsequent layer into one function.

    \hfill\\
    \ptitle{ToyShapes} 
    We generate synthetic datasets consisting of images divided into quadrants, within each of which a geometric shape may be drawn (square, triangle, or circle). Heterogeneity is introduced in the shades of gray for both the background and the shapes, as well as in the size of the shapes, and we let one quarter of the shapes be non-filled (outlined). Examples can be found in Fig. \ref{fig:toyshapes_detection}. 

    We consider a binary classification task where the positive class corresponds to the presence of at least one triangle in the image. A single logit $\hat y$ is constructed and the predicted class is obtained through a simple thresholding operation: \enquote{triangle(s)} if $\hat y>0$ (equivalent to Sigmoid$(\hat y)>0.5$). We set $\Z\equiv\X$, meaning that we construct explanations in the input space. Explanations take the form of attribution maps $\pibh\in\Z$ or $\pib\in\Z$ and are expected to detect the pixels shaping triangles. 

    In PiNets, attribution maps $\pib$ are produced by the decoder. In the ToyShapes settings, we let the decoder culminate in a sigmoid transformation such that $\hat\pi_d(\x)\in[0,1],\ \forall d$. We grant the model additional flexibility by allowing it to learn both a global intercept $\hat a$ and a global scale parameter $\hat b$. We constrain $g(\pib,\z)$ to be increasing in the coefficients by squaring the scaling parameter. This simplifies interpretation across models: coefficients close to 0 always translate into low pixel importance and vice versa. The resulting binary PiNet classifier takes the form:
    \begin{equation}
      \hat{y} = \hat a + \hat b^2 \sum\nolimits_* \pib \circ \z
      \label{eq:binary_pinet}
    \end{equation}

    \hfill\\
    \ptitle{Experimental settings for ToyShapes} 
    We compute Grad-CAMs \citep{GRADCAM} over convolutional neural networks (CNNs \citep{CNN}) by way of baseline explanations. Grad-CAMs are known to deliver state-of-the-art feature attributions in image classification tasks and pass fundamental sanity checks \citep{GRADEVAL1}. The CNN architecture of the baseline is the same as the encoder of the PiNets we evaluate. Alongside the baseline approach, six PiNet variants are considered:
    
    \needspace{3cm}
    \begin{itemize}[itemsep=-3pt, topsep=3pt]
      \item[{\fdot[Ccam]{1.5}}] \textbf{Grad-CAM} — CNN + Grad-CAM
      \item[{\fdot[Cnaive]{1.5}}] \textbf{PiNet Naive} — PiNet with inadequate decoder
      \item[{\fdot[Csoft]{1.5}}] \textbf{PiNet Soft} — PiNet with soft second look
      \item[{\fdot[Cpinet]{1.5}}] \textbf{PiNet} — default PiNet
      \item[{\fdot[Cfeedback]{1.5}}] \textbf{PiNet Feedback} — PiNet with recursive feedback
      \item[{\fdot[Censemble]{1.5}}] \textbf{PiNet Ensemble} — ensemble of PiNets
      \item[{\fdot[Cstrong]{1.5}}] \textbf{PiNet Strong} — strongly-supervised PiNet 
    \end{itemize}
    
    The default PiNet \fdot[Cpinet]{1.5} uses an adequate decoder composed of transposed convolutions (akin to fully convolutional networks \citep{FCN}), a hard second look, no recursive feedback, and no strong supervision. 

    To probe results stability, we train 30 CNNs for the baseline and 30 PiNets of each variant, as well as 30 ensembles composed of 10 default PiNets each. Both the training data (1,000 examples) and model parameters are reinitialized at each run with a different random seed. Each time, a 20\% validation set is held out from the training data and the model is trained for up to 50 epochs unless validation accuracy reaches 100\% earlier. Otherwise, the last model checkpoint is saved only if validation accuracy is greater than or equal to 98\%. Strongly-supervised PiNets receive 25 ground-truth attribution maps $\pibs$ in addition to the 800 class labels, corresponding to exactly one map per step within each training epoch (the batch size is 32). The other hyperparameters governing model architectures and the training procedure can be found in the code repository.

    By construction, PiNets' coefficients lie in $[0,1]$. For Grad-CAMs, we obtain the best results by zeroing out negative gradients, as is common practice. Subsequently, and for both approaches, attribution maps are normalized in $[0,1]$ through a division by the maximum coefficient or gradient observed on the test set, composed of 1,000 held-out examples (different from the validation sets).

    We consider two approaches to post-processing attribution maps when producing the final explanations. The procedure described above yields continuous attributions $\hat\pi_d(\x)\in[0,1],\ \forall d$. In contrast, fine-tuning binarizes the attributions using a threshold $t\in[0,1]$ selected to maximize a detection score reflecting meaningfulness (eq. \eqref{eq:score}). This second approach yields binary attributions $\hat\pi_d(\x)\in\set{0,1},\ \forall d$. 
    
    Threshold fine-tuning is performed on the test set. Doing it on training data would suggest that ground-truth training attributions are available and can be used to train the model, which we do not assume except for the strongly supervised PiNets. Fine-tuning thus presupposes that ground-truth test attributions are available for meaningfulness quantification. We assessed the difficulty of fine-tuning attributions precisely to address this common practical limitation (in Results).

    \hfill\\
    \ptitle{Meaningfulness} 
    Explanations are attribution maps $\pibh:\X\to\Z$ (or $\pib$) and we gauge their meaningfulness through their attribution accuracy with respect to binary ground-truth attributions $\pibs$ (positive pixels delineate triangles). Analogous to performance indicators used in binary classification, we distinguish between two types of errors. The {True Detection Rate} (TDR) parallels the true positive rate (\textit{aka} sensitivity or recall); its complement represents the miss rate. A low TDR indicates that the method fails to detect the relevant signal. The {True Abstraction Rate} (TAR) parallels the true negative rate (\textit{aka} specificity); its complement represents the false alarm rate. A low TAR indicates that the method fails to filter out irrelevant or spurious signal. 
    
    To accommodate both continuous and binary attribution maps, we define and combine the TDR and TAR for any instance $\x\in\X$ as follows (applies to $\pib$ equivalently):

    \begin{align}
      &\quad\text{TDR} = \frac{\sum_* \pibs \circ \pibh}{\sum_* \pibs} \label{eq:tdr} \\
      &\quad\text{TAR} = \frac{\sum_* \neg\pibs \circ \neg\pibh}{\sum_* \neg\pibs} \label{eq:tar} \\
      &\textbf{Detection Score} = \overline{\text{TDR}} \times \overline{\text{TAR}} \label{eq:score}
    \end{align}
    
    Where the operator $\neg$ denotes the complement ($1-\cdot$) of the object it precedes; that is, $\neg\pib$ is the negative image of $\pib$. Instance-wise TDRs and TARs are averaged over the test set ($\overline{\text{TDR}}, \overline{\text{TAR}}$), and the detection score is the product of these averages. This composite indicator rewards balance between sensitivity and specificity, and carries a useful interpretation: for any value $\alpha\in[0,1]$ taken by the detection score, both components ($\overline{\text{TDR}}$ and $\overline{\text{TAR}}$) are guaranteed to be greater than or equal to $\alpha$, since they are both in $[0,1]$.

    \hfill\\
    \ptitle{Sufficiency and Robustness} 
    We compute the accuracy shift under recursive prediction; that is, the change in predictive accuracy when predicting from the filtered signal $\pib\circ\z$ (or $\pibh\circ\z$) compared to the original signal $\x$. The recursive prediction is trivial to perform when $\Z=\X$ or when they coincide in structure. More generally, there must be a way to go from $\Z$ to $\X$ so as to transform $\pib\circ\z$ into $\pib\circ\x$.
    
    To be sufficient, an explanation must detect signals so as to recover the prediction if predicting from it. A lack of sufficiency would be reflected in distorted recursive predictions and thus in a change in predictive accuracy (most likely a drop). 
    
    Besides, a lack of robustness to context can be responsible for discrepancy between the initial and the recursive explanations. Because in PiNets this implies recursive instability of predictions, robustness can also be captured by recursive accuracy shifts.

    \vspace{.5cm}
    \begin{figcol}
      \centering
      \includegraphics[width=1\textwidth]{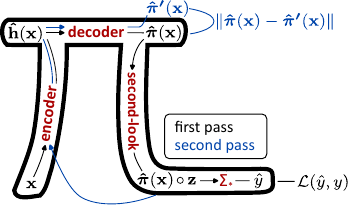}
      \captionof{figure}{PiNet with recursive feedback. The explanation is used to construct the recursive input $\pib\circ\z$. The discrepancy between the initial explanation $\pib$ and the recursive explanation $\piboldhat'(\x)$ is penalized.}
      \label{fig:pinet_feedback}
    \end{figcol} 

    \hfill\\
    \ptitle{Recursive stabilization} 
    As a direct response to the criteria of robustness to context and sufficiency we propose a training technique that targets the {recursive stability} of explanations. As illustrated in Fig. \ref{fig:pinet_feedback}, the loss function is augmented with a feedback on the dissimilarity between the initial explanation $\pib$ and the one generated recursively from $\pib\circ\z$ as input (feasible in this context since $\Z\equiv\X$).
    
    Specifically, we penalize the discrepancy between the initial explanation $\pib$ and the recursive explanation $\piboldhat'(\x)$ (when $\X\equiv\Z$ we have $\piboldhat'(\x)=\piboldhat(\pib\circ\z)$). We consider a norm distance by way of feedback loss:
    \begin{equation}
      \textbf{Feedback loss} = \left\| \pib - \piboldhat'(\x) \right\|
      \label{eq:feedback}
    \end{equation}

    Minimizing the feedback loss directly pushes model explanations to be recursively stable. Then, if attributions $\pib$ filter signal out, the sparser recursive input $\pib\circ\x$ should be exploited by the model to build similar attributions. This implies recursive stability in explanations and thus robustness to context; and since predictions are directly built from explanations, it also implies recursive stability in predictions and thus sufficiency.

    \hfill\\
    \ptitle{Ensembling} 
    Leveraging multiple models through ensembles is a common strategy to improve predictive accuracy \citep{MOE,STACKING,NETENSEMBLE,BAGGING,FOREST}. Ensembling can smooth out the errors of each component model, hence improving generalization accuracy. However, it may be at odds with explainability \citep{ENSEMBLECOST1,ENSEMBLECOST2,ENSEMBLECOST3}. A typical example is random forests \citep{FOREST}, which blend multiple reasonably readable decision trees into an ensemble that becomes painstaking to read.

    An ensemble of neural networks risks scoring low on explainability. But additively ensembling PiNets comes down to linearly combining pseudo-linear models, an operation that preserves pseudo-linearity and thus readability. Furthermore, akin to the improvement of predictive accuracy through averaging, model-specific explanation errors could also be evened out in the ensemble, which is largely confirmed by our results.

    Assuming that the aggregator is a summation and that the ensemble is uniformly weighted, we can write an ensemble of $M$ pseudo-linear models as follows:

    \begin{align}
      \hat{\underline y} &= \frac{1}{M} \sum_m \hat y_m 
        = \frac{1}{M} \sum_m \parent{ \hat a_m + \sum\nolimits_* \piboldhat_m(\x) \circ \z } \nonumber \\
        \implies \hat{\underline y} &= \parent{\sum_m \frac{\hat a_m}{M}} + \sum\nolimits_* \parent{\sum_m \frac{\piboldhat_m(\x)}{M}} \circ \z \nonumber \\
        \implies \hat{\underline y} &= \hat{\underline a} + \sum\nolimits_* \boldsymbol{\hat{\underline\pi}}(\x) \circ \z
    \end{align}

    Notice how the ensemble collapses into a single model with aggregated coefficients $\boldsymbol{\hat{\underline\pi}}(\x)$ and shift parameter $\hat{\underline a}$, effectively preserving readability and alignment. 
    
    If scale parameters $\{\hat b_m\}_m$ are included, they end up weighing the attribution maps of individual PiNets in the ensemble. For example, we combined $M=10$ binary PiNet classifiers in the ToyShapes experiments, which yields the aggregate attribution map: 
    \begin{equation}
      \boldsymbol{\hat{\underline\pi}}(\x) = \frac{1}{10} \sum_{m=1}^{10} \hat b_m^2 \cdot \piboldhat_m(\x)
      \label{eq:binary_pinet_ens_det} 
    \end{equation}

    \hfill\\
    \ptitle{Strong supervision} 
    When available, ground-truth attributions $\pibs$ can be used to supervise the training of PiNets. To this end, the training dataset is augmented with ground-truth explanations and an attribution loss capturing the meaningfulness (accuracy) of model explanations $\pib$ is added to the prediction loss used to supervise training. Depending on the context, this can be a cross-entropy loss, a Dice loss \citep{DICE1,DICE2}, or a distance-based loss. In the latter case we write: 
    \begin{equation}
      \textbf{Attribution loss} = \left\| \pib - \pibs \right\|
      \label{eq:attribution_loss}
    \end{equation}

    \hfill\\
    \ptitle{Experimental settings for flood mapping}
    The detection of flooded areas typically requires the availability of high-quality hand-annotated segmentation maps as proxies for ground-truth $\pibs$. Segmentation models, usually taking the form of encoder-decoder architectures, can be trained to classify each pixel \citep{FCN,UNET}. We follow this approach to train a baseline model that we call SegNet. 
    
    In addition, we train a PiNet to predict the surface area of flooded and non-flooded regions. For this purpose, response variables $\y=\set{\hat y_{-1},\hat y_0,\hat y_1}$ are constructed to represent the number of pixels occupied by each class (not valid, no water, and water, respectively). The regression is therefore performed based on image-level information, whereas the SegNet is exposed to ground-truth attribution maps that embed pixel-level information. In this respect the SegNet can be seen as a PiNet model exclusively strongly supervised (on $\pibs$), whereas the PiNet is exclusively weakly supervised (on $\mathbf y$). The PiNet is equipped with a soft second look and is, in fact, equivalent to a SegNet augmented with an aggregator enabling regression. The same encoder-decoder architecture is shared by both the SegNet and the PiNet, as well as training settings. The backbone encoder is the geospatial foundation model Prithvi \citep{PRITHVI, TERRATORCH}, while the decoder is a UPerNet \citep{UPERNET}. Only the decoder is trained as we freeze the pretrained weights of the encoder. 
    
    We let $\Z$ be a matrix space whose dimensions correspond to the height and width of the satellite images, while the input space $\X$ has six bands of same dimensions. For any pixel $d$, a soft classification is produced from a softmax normalization of the decoder's output, yielding $\hat\pi_{d}\in[0,1]^3$ such that $\sum_k\hat\pi_{d,k}=1$ and $\sum_d\sum_k\hat\pi_{d,k}=|\Z|$, i.e., the number of pixels in the image. Then, PiNets' predictions take the form $\hat y_k=\sum_* \pibk,\ \forall k$. In the final segmentation maps used for visualization (Fig. \ref{fig:floods}), each pixel is assigned the class with highest probability.   

    Referring to Table \ref{tab:floods}: for the water and no-water classes we report the true detection rate (TDR) defined in eq. \eqref{eq:tdr} and the intersection-over-union (IoU), which is defined for each class $k$ as:
    \begin{equation}
      \text{IoU}_k = \frac{ \sum_* \pibk \circ \pibold_k(\x) } 
                    { \parent{\sum_* \pibk + \pibold_k(\x)} - \parent{\sum_* \pibk \circ \pibold_k(\x)} }
    \end{equation}  

    In addition, the mean absolute error (MAE), used as loss function to train the PiNet, is reported for the water class. It corresponds to the test-set average of the water absolute error (AE$_1$) defined as:
    \begin{equation}
      \text{AE}_1(\x) = \sum_*\ | \piboldhat_1(\x) - \pibold_1(\x) |
    \end{equation}  

    The regression performance is not of great importance here. If we were interested in  estimating flooded surface areas, then the MAE of the water class would be a key indicator, whereas the MAE for the no-water and not-valid classes would be of little relevance.

    \hfill\\
    \ptitle{Implementation and reproducibility}
    Complete code for the generation of the synthetic data and the reproduction of the results is available in the Github repository: \url{https://github.com/FractalySyn/PiNets-Alignment}. We used PyTorch for the implementation and training of deep learning models \citep{PYTORCH} and the Adam algorithm \citep{ADAM, ADAMW} for optimization.

    \hfill\\
    \ptitle{Acknowledgments} We are thankful to S. Bianchini (Unistra), B. Gulhan (Penn State), M. Laguerre (EM Lyon), R. Porcedda (Sant'Anna), and L. Testa (Carnegie Mellon) for their insightful comments and suggestions on the manuscript.

\end{multicols}

%%%%%%%%%%%%%%%%%%%%%%%%%%%%%%%%%%%%%%%%%%%%%%%%%%%%%%%%%%%%%
%%%%%%%%%%                 Bottom                 %%%%%%%%%%%
%%%%%%%%%%%%%%%%%%%%%%%%%%%%%%%%%%%%%%%%%%%%%%%%%%%%%%%%%%%%%

% \newpage
% \hfill\\
\needspace{3cm}
\small
\begin{multicols}{3}
  \bibliographystyle{mystyle.bst} % unsrtnat + abbrvnat
  \bibliography{references} 
\end{multicols}

%%%%%%%%%%%%%%%%
\end{document}